\documentclass[final]{cvpr}

\usepackage{times}
\usepackage{epsfig}
\usepackage{graphicx}
\usepackage{amsmath}
\usepackage{amssymb}
\usepackage{tabularx}
\usepackage{multirow}
\newcommand{\eqnref}[1]{Eq.~(\ref{#1})}

\usepackage[pagebackref=true,breaklinks=true,colorlinks,bookmarks=false]{hyperref}

\def\cvprPaperID{10892} 
\def\confYear{CVPR 2021}

\begin{document}

\title{Depth Completion using Plane-Residual Representation}

\author{Byeong-Uk Lee \quad\quad Kyunghyun Lee \quad\quad In So Kweon\\
Korea Advanced Institute of Science and Technology\\
Daejeon, Republic of Korea\\
{\tt\small \{byeonguk.lee, kyunghyun.lee, iskweon77\}@kaist.ac.kr}
}

\maketitle
\thispagestyle{empty}
\pagestyle{empty}

\begin{abstract}
The basic framework of depth completion is to predict a pixel-wise dense depth map using very sparse input data.
In this paper, we try to solve this problem in a more effective way, by reformulating the regression-based depth estimation problem into a combination of depth plane classification and residual regression.
Our proposed approach is to initially densify sparse depth information by figuring out which plane a pixel should lie among a number of discretized depth planes, and then calculate the final depth value by predicting the distance from the specified plane.
This will help the network to lessen the burden of directly regressing the absolute depth information from none, and to effectively obtain more accurate depth prediction result with less computation power and inference time.
To do so, we firstly introduce a novel way of interpreting depth information with the closest depth plane label $p$ and a residual value $r$, as we call it, Plane-Residual (PR) representation.
We also propose a depth completion network utilizing PR representation consisting of a shared encoder and two decoders, where one classifies the pixel's depth plane label, while the other one regresses the normalized distance from the classified depth plane.
By interpreting depth information in PR representation and using our corresponding depth completion network, we were able to acquire improved depth completion performance with faster computation, compared to previous approaches.
\end{abstract}

\section{Introduction}
\label{sec:intro}

Many different computer vision algorithms are becoming more reachable in our everyday life, starting from a smartphone camera and augmented reality (AR) / virtual reality (VR) applications to autonomous driving and even more complicated robotics tasks.
In order to solve these problems efficiently, obtaining precise and reliable 3D scene information is crucial.
3D reconstruction has been studied for many years and certainly can be categorized as one of the traditional computer vision tasks, but it is still a core technology.
There are various ways of obtaining 3D information, such as monocular depth prediction, structure from motion, and multi-view stereo.
However, 3D reconstruction using additional sensors alongside an RGB camera like secondary camera, depth sensors, or radar will be more effective, since it can utilize more accurate depth measurements as prior information.
Given that most of the recent mobile devices have more than one camera and even a LiDAR sensor, it is widely acceptable that 3D reconstruction by integrating multiple sensor inputs is a more efficient and practical approach.

While stereo matching being one of the most conventional and reliable ways in 3D reconstruction, it shows its weakness in depth prediction on the farther area due to the physical restriction for large baseline between stereo cameras.
Therefore, using a depth sensor to obtain more accurate and absolute initial depth information is also preferred.
However, one of the major downsides of commercialized depth sensors, such as 3D LiDAR, Kinect, and RealSense, is the sparsity of the measurement.
Addressing this problem, various approaches emerged trying to densify, i.e., `complete' the sparse depth measurement into a dense depth map, namely, `depth completion'.

In recent years, there have been many different methods trying to solve depth completion using deep learning, starting from Ma and Karaman~\cite{ma2018s2d}.
The challenging part of these algorithms is that regression-based approaches have difficulties in maintaining the information of the object boundary and may show mixed depth results~\cite{imran2019dc}.
A few early works addressed these difficulties by maximizing the information from the RGB input and refining the initial depth regression output to get better final results~\cite{cheng2018cspn, ma2019self, lee2019context, van2019sparse, cheng2020cspn, cheng2020cspn++, park2020nlspn}.
Other algorithms tried to utilize additional information that can be inferred from the depth map, such as surface normal, to give more geometrical guidance to the training process~\cite{lee2019context, xu2019depth, qui2019deeplidar}.
While these algorithms showed some promising outcomes, they still lack in preserving edge information and often require a large amount of computation power, as in heavy network memory and longer inference time, which are not suitable for real-time applications.

To tackle these problems, we introduce a Plane-Residual (PR) representation, a novel way of interpreting depth information, and an end-to-end depth completion deep learning network.
PR representation expresses an absolute depth value of a pixel with two parameters $(p, r)$, where $p$ refers to the closest depth plane among a number of pre-defined discretized depth planes, and $r$ refers to the normalized distance from the selected plane.
With Plane-Residual representation, we can factorize the direct depth regression problem into a combination of discrete depth plane classification and plane-by-plane residual regression.
We also propose an end-to-end depth completion network using PR representation to execute this idea and present our results with good performance and fast computation.
We did extensive ablation studies and compared our algorithm with some state-of-the-art methods, both quantitatively and qualitatively, to prove the effectiveness and the validity of our design choices.

\section{Related Work}
\label{sec:rw}

\subsection{Depth Completion}

Ma and Karaman~\cite{ma2018s2d} proposed depth completion using convolutional neural network (CNN).
They utilized the overall network structure from Laina \etal~\cite{laina2016deeper} to solve the depth completion problem using an RGB image and a single-channeled sparse depth input.
Their algorithm showed that using only a fraction of additional input depth information boosts a big margin of performance.

Cheng \etal~\cite{cheng2018cspn} and Park \etal~\cite{park2020nlspn} addressed the blurriness of the result in depth regression using deep learning, and tried to solve this problem by utilizing post-processing refinement module called convolutional spatial propagation network (CSPN).
CSPN learns affinity weights of each pixel to its neighbor pixels, where those weights are used to refine initial depth result iteratively.
Park \etal~\cite{park2020nlspn} compensated CSPN by proposing a non-local spatial propagation network (NLSPN), to learn non-local affinity weights that are not restricted to a square-shaped propagation kernel.
There are more approaches using different versions of CSPN~\cite{cheng2020cspn, cheng2020cspn++}, but these works all suffer from slow inference time.

Imran \etal~\cite{imran2019dc} approached in a slightly different way, by transforming the depth information into a series of depth coefficient for multiple discrete depth basis.
They tried to address the problem that deep learning based depth regression using upsampling might lose the information of the object boundaries.
By doing so, they were able to propose relatively light-weighted depth completion network, but lacked on the performance.

Chen \etal~\cite{chen2019fusenet} introduced a way to maximize the 3D information from the sparse depth input.
They applied 2D convolution on image pixels and continuous 3D convolution on point cloud, so that each branch extracts useful features from each side.
They fused the features from both sides in a multi-scaled manner, to get more accurate depth completion result, with well-preserved object boundary information.





\subsection{Depth Prediction using Multiple Planes}

Imran \etal~\cite{imran2019dc} used discretized multiple planes and corresponding coefficients. 
They represented the depth value of each pixel by a weighted summation of pre-defined discretized depth planes, and learned those coefficients by multi-label classification. 

Fu \etal~\cite{fu2018deep} proposed the ordinal regression to estimate the depth from a monocular image. 
They compared multi-label classification and ordinal regression with predefined multiple depth planes. 
The proposed deep ordinal regression network (DORN) divided the depth estimation problem into sub-problems of plane-wise binary classification, each determining whether the true depth value is larger than the discrete depth of each plane. 
However, they used the discrete depth value itself in the final depth generation, therefore, the accuracy is limited. 

Meanwhile, plane-sweeping algorithms are famous stereo matching methods using multiple planes. \cite{flynn2016deepstereo, huang2018deepmvs, im2018dpsnet}
They utilize cost-volume by stacking the photometric loss of the warped image from every discrete plane. 
Im \etal~\cite{im2018dpsnet} integrated the process into the deep learning framework and proposed an end-to-end plane-sweeping algorithm. 
However, they need a new image from a different viewpoint with known extrinsic parameters to calculate cost volume.
Therefore, they are usually used in a stereo setup or with consecutive image sequences. 

Other methods to represent an image with multiple planes are Multi-Plane Image (MPI). 
Tucker \etal~\cite{tucker2020mpi} and Zhou \etal~\cite{zhou2018stereo} represented a single RGB image as multiple RGBA images. 
In this kind of representation method, the final image is obtained by the cumulative multiplication of the layers with additional transparency value. 
However, these methods are not suitable to be directly used in depth prediction, since they were designed for differentiable rendering and view synthesis.

\section{Methodology}
\label{sec:meth}
In this section, we elaborate our new way of representing depth information, which we named Plane-Residual (PR) Representation, with detailed network structure and design choices for our end-to-end depth completion algorithm.
Overall pipeline of our approach is shown in~\figref{fig:pipeline}.

\subsection{Plane-Residual Representation}
\label{sec:pr}

\begin{figure*}[t]
\begin{center}
\includegraphics[width=\linewidth]{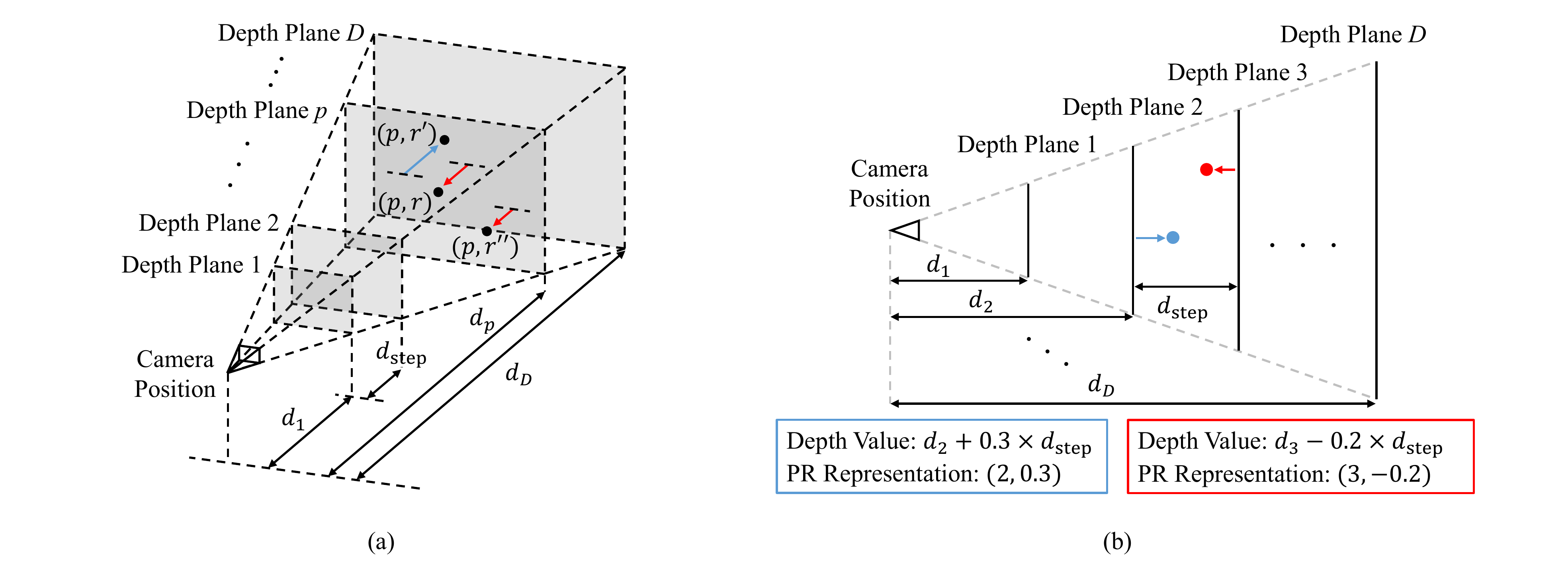}
\end{center}
\caption{\textbf{PR representation.} (a) 3D illustration of our PR representation. (b) 2D illustration and examples of how absolute depth values are interpreted in PR representation. Equations inside the blue-lined and red-lined boxes notate the PR representations from the actual depth values of the blue dot and the red dot, respectively.
}
\label{fig:pr}
\vspace{-2mm}
\end{figure*}

As mentioned in~\secref{sec:intro} and~\secref{sec:rw}, directly estimating depth information with deep learning using regression is challenging.
Therefore, we try to solve depth estimation by factorizing a challenging regression problem into a combination of relatively simple classification and regression.

To enable this idea, first we introduce a Plane-Residual (PR) representation, a novel way of interpreting depth information.
Our PR representation assumes a set of $D$ fronto-parallel planes to the camera, where each plane represents a discrete depth value $d_{p\in\{1, 2, \cdots, D\}}$.
Depth plane 1 is the closest plane from the camera, and depth plane $D$ is the farthest.
The values of $d_1, d_2, \cdots, d_D$ and $d_{\text{step}}$, which is the distance between two adjacent depth planes, can be determined arbitrarily.

Within a depth image, the depth value of each pixel is then expressed as $(p, r)$, where $p$ refers to the closest depth plane that the pixel lies.
$r$ is the normalized distance, \ie, \textit{residual} from the plane depth to its actual value according to $d_{\text{step}}$.
Therefore, every $r$ values are in range of $[-0.5, 0.5)$, except $[0, 0.5)$ for depth plane 1 and $[-0.5, 0]$ for depth plane $D$.
Given the values of the plane depths $d_p$s, the absolute depth value of a pixel represented as $(p, r)$ is,
\begin{equation}
\label{eq:pr}
\begin{split}
    D(p, r) & = d_p + r \times d_{\text{step}}(p, r), \\
    d_{\text{step}}(p, r) & = 
    \begin{cases}
        d_{p+1} - d_p & \text{if} \; r \geq 0 \\
        d_p - d_{p-1} & \text{if} \; r < 0 \\
    \end{cases}
\end{split}
\end{equation}
More intuitive illustration of our PR representation is shown in~\figref{fig:pr}.

By using PR representation, we can seamlessly reformulate the depth completion task into a combination of depth plane classification and relative residual regression.
Depth plane classification simplifies the densification of the given sparse depth information. 
Relative residual regression makes the final depth result to be more accurate and reduces discontinuity issue from the discretized depth planes.

\subsection{Network Design}
\label{sec:net}

\begin{figure*}[t]
\begin{center}
\includegraphics[width=\linewidth]{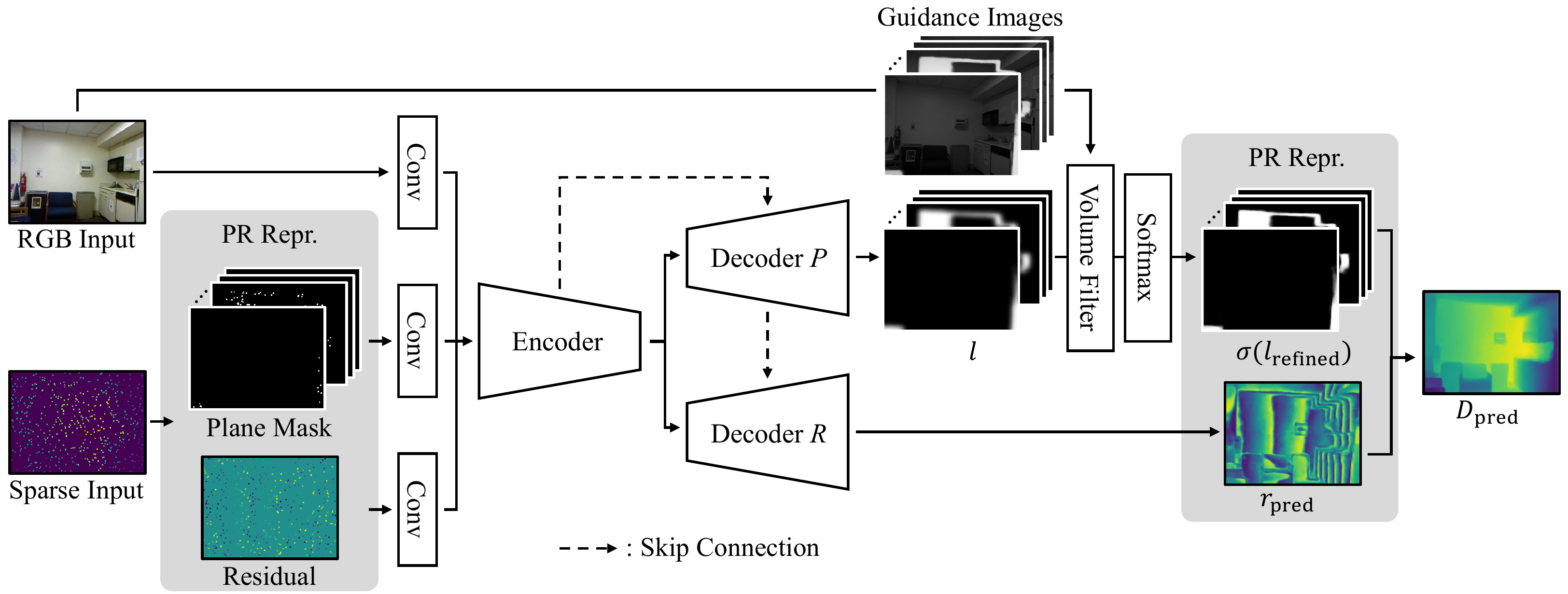}
\end{center}
\caption{\textbf{Overall pipeline of our depth completion algorithm.} With a single RGB image and a sparse depth input, we solve depth completion with a combination of depth plane classification and residual regression. Please refer to~\secref{sec:meth} for detailed information.}
\label{fig:pipeline}
\vspace{-2mm}
\end{figure*}

\begin{figure}[t]
\begin{center}
\includegraphics[width=\linewidth]{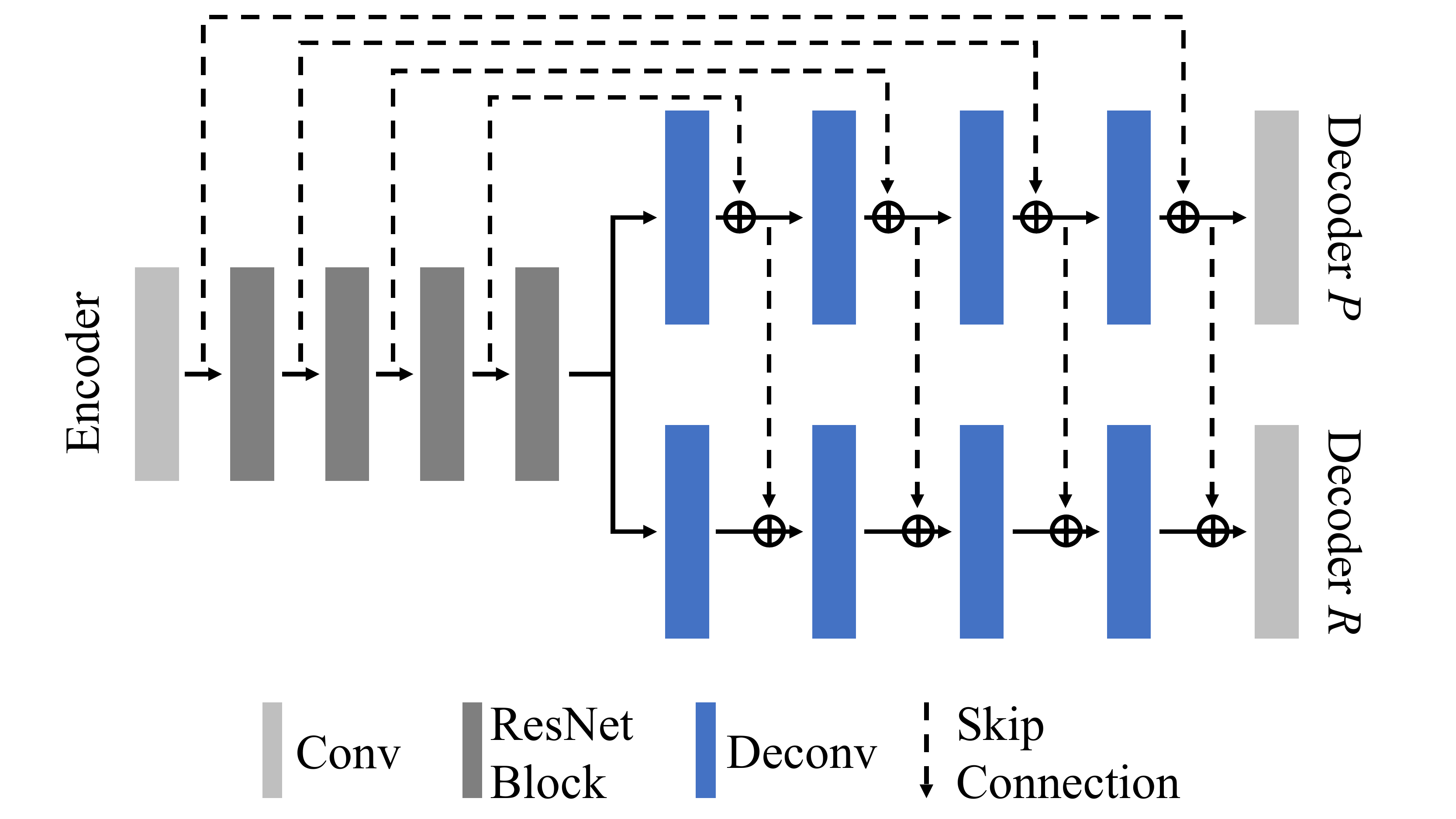}
\end{center}
\caption{\textbf{Illustration of the detailed network architecture and our late feature fusion technique.} Decoded feature maps from decoder $P$ are fused to the decoded feature maps in decoder $R$, to guide residual regression with plane classification information.}
\label{fig:detail}
\vspace{-2mm}
\end{figure}

Our depth completion network using PR representation consists of a single ResNet-based~\cite{he2016resnet} encoder and two decoders, decoder $P$ and decoder $R$.
The overall pipeline of our approach is shown in~\figref{fig:pipeline}.

The encoder takes an RGB image and sparse depth information as an input.
Input sparse depth is pre-processed into a PR representation.
The P part of the PR representation is expressed in a $D$-channeled valid mask, and the R part is expressed in a single-channeled residual map.
Each type of input element is processed with a single convolution layer and then concatenated all together before the ResNet-based encoder~\cite{ma2019self, cheng2020cspn++}.

Two decoders are similar structure-wise while having distinguished roles.
Each decoder block in both decoders consists of a deconvolution~\cite{noh2015learning} layer.
Skip connection~\cite{ronneberger2015unet} from the encoder is given to every matching output of the decoding blocks in decoder $P$, which is known to be effective for retaining object boundary information.
Moreover, in our proposed PR representation, residual value $r$ is determined in regard of the depth plane $p$.
Therefore, some kind of guidance from plane classification process should be given when regressing residual map.
To give this guidance while maintaining the parameter-wise effectiveness of the whole depth completion network, we chose to give additional skip connection from the decoded features of decoder $P$ into the decoded features of decoder $R$.
More detailed and intuitive architecture of our depth completion network using PR representation is shown in~\figref{fig:detail}.
We chose feature summation rather than feature concatenation for our skip connection, since encoded features are in the similar domain as decoded features~\cite{qui2019deeplidar}.

Decoder $P$ outputs a $D$-channeled probability volume, using a softmax function.
The output of the decoder $R$ is a single-channeled normalized residual map, $r_{\text{pred}}$.
Using~\eqnref{eq:pr}, the final prediction of the network $D_{\text{pred}}$ is acquired by,
\begin{equation}
\begin{split}
    D_{\text{pred}}(x, y) = & \sum_{p = 1}^{D} d_p \times \sigma(l_p(x, y)) \\
    & + r_{\text{pred}}(x, y) \times d_{\text{step}}(\hat{p}, r_{\text{pred}}(x, y)).
    \label{eq:pred}
    \end{split}
\end{equation}
Here, $\sigma ( \cdot )$ is a softmax function, and $l_p(x, y)$ is the raw classification logit at the coordinate $(x, y)$ for depth plane $p$.
$\hat{p}$ is the classified depth plane at the pixel $(x, y)$ with the highest probability of $\sigma(l_p(x, y))$.
Unlike~\eqnref{eq:pr}, we used weighted summation for the final depth estimation with the probability volume $\sigma (l_p)$, to smoothen the depth result in the boundary area of two discrete depth planes.

\subsection{Probability Volume Filtering}
\label{sec:filter}
In order to achieve more accurate depth plane classification results, we apply channel-wise guided image filtering.
Guided image filtering~\cite{he2010gf, he2013gf} uses content information from the guidance image to perform edge-preserving smoothing.
Applying this technique to each channel of the cost volume for correspondence matching was introduced and was shown to be effective in~\cite{hosni2013filter}.
Since our probability volume for depth plane classification has similar structure to the cost volume in plane-sweeping and stereo matching, we believe that this kind of technique can be easily applied to our approach as well.

To adopt this idea, we first apply two consecutive convolution layers to the input RGB image to create $D$-channeled guidance images $I_{\text{guide}}$, as in~\cite{wu2017fast}.
Then we perform channel-wise guided image filtering to the initial depth plane classification logit $l$, by
\begin{equation}
    l_{\text{refined}, p}^i = A_p^k I_{\text{guide}, p}^i + b_p^k, \; \forall i \in w_k,
\label{eq:gf}
\end{equation}
where $i$ is a pixel location, $w_k$ is a window centered at the pixel $k$, and $l_{\text{refined}, p}$ is a refined classification logit at depth plane $p$.
$A_p^k$ and $b_p^k$ are determined using mean, variance and covariance values of $I_{\text{guide}, p}$ and $l_p$ in a window $w_k$, where we use an average pooling layer to utilize this equation in the training process.
Our final depth result will be modified from~\eqnref{eq:pred} by substituting $l$ to $l_{\text{refined}}$, as in,
\begin{equation}
\begin{split}
    D_{\text{pred}}(x, y) = & \sum_{p = 1}^{D} d_p \times \sigma(l_{\text{refined}, p}(x, y)) \\
    & + r_{\text{pred}}(x, y) \times d_{\text{step}}(\hat{p}, r_{\text{pred}}(x, y)).
    \label{eq:refined}
    \end{split}
\end{equation}

\subsection{Confidence-Based Regression}
\label{sec:cf}

As described in~\secref{sec:net}, the residual value $r$ depends on the classified depth plane $p$ of a pixel.
Therefore, reliable depth plane classification performance is required for robust residual learning.
If the decoder $R$ was trained with supervision of the ground truth residual map while the decoder $P$ gives false depth plane prediction, the final depth map will result in poor depth map quality with discretizing artifacts on the depth plane boundary region.

To address this problem, we measure depth plane classification confidence to use it as a training loss weight for decoder $R$.
The core idea is to give the network optimizer some guidance on where to put heavier weight between ground truth depth map $D_{\text{gt}}$, or ground truth residual map $r_{\text{gt}}$.
For confidence measure, we use the channel-wise maximum value of the probability volume $\sigma(l)$ on each pixel, as in,
\begin{equation}
    c(x, y) = \max(\sigma(l_1(x, y)), \cdots, \sigma(l_D(x, y))),
\label{eq:cf}
\end{equation}
where $c(x, y)$ is the confidence value at the pixel coordinate $(x, y)$.

For a pixel with high depth plane classification confidence, more supervision of the ground truth residual value is given, and for a pixel with low depth plane classification confidence, the decoder $R$ will be trained with more guidance from $D_{\text{gt}}$.

In the earlier stage of the training, the network can focus more on the depth plane classification.
In the latter stage of the training, when the depth plane classification becomes more reliable, the network can learn residual map more efficiently.
Also, for the boundary area of two adjacent depth planes, the decoder $R$ will be trained adaptively between the ground truth residual map and the ground truth depth map.

\subsection{Loss Function}

For accurate depth prediction and effective training, we trained our depth completion network with three loss terms, which are final depth loss, depth plane classification loss, and residual loss.

\noindent{\textbf{Final depth loss.}} The first loss term is a simple $L1$ loss between the final depth prediction $D_{\text{pred}}$ from~\eqnref{eq:refined} and the ground truth depth $D_{\text{gt}}$, as in,
\begin{equation}
    L_D = |D_{\text{gt}} - D_{\text{pred}}|.
\end{equation}

\noindent{\textbf{Depth plane classification loss.}} To maximize the benefits of our PR representation, we apply cross-entropy loss for the depth plane classification from decoder $P$.
Therefore, the cross-entropy loss $H(P_{\text{gt}}, l)$ for our depth plane classification logit $l$ can be described as,
\begin{equation}
    H(P_{\text{gt}}, l) = -\sum_{p=1}^D P_{\text{gt}}(p) \log ( \sigma (l_p)).
\end{equation}
$P_{\text{gt}}(p)$ is a ground truth depth plane classification binary indicator, showing if the pixel is in the depth plane $p$.

As described in~\secref{sec:filter}, we also have a refined logit output $l_{\text{refined}}$.
We calculate the same loss for both initial and refined classification prediction.
Our final depth plane classification loss $L_P$ will be defined as following:
\begin{equation}
    L_P = \lambda H(P_{\text{gt}}, l) + H(P_{\text{gt}}, l_{\text{refined}}),
\end{equation}
with $\lambda$ being a loss weight parameter which was set to 0.7, empirically.

\noindent{\textbf{Residual loss.}} The last loss term is also a simple $L1$ loss, calculated between the residual prediction $r_{\text{pred}}$ and the ground truth residual map $r_{\text{gt}}$.
With the confidence explained in~\secref{sec:cf}, our residual loss can be interpreted as,
\begin{equation}
    L_R = c \odot |r_{\text{gt}} - r_{\text{pred}}|,
\end{equation}
where $\odot$ indicates the Hadamard product.

Our final loss function $L$ for end-to-end training of our depth completion network is,
\begin{equation}
    L = L_D + L_P + \frac{1}{D}L_R.
\end{equation}
Since the residual part of our PR representation is normalized to a value of $d_{\text{step}}$, we penalize the residual loss with the parameter of $\frac{1}{D}$.

\section{Experiment}
\label{sec:exp}

\subsection{Dataset \& Training}

\noindent{\textbf{NYU Depth V2 dataset.}} The NYU Depth V2 dataset~\cite{silberman2012nyuv2} contains RGB and depth sequences in 464 indoor scenes.
The dataset was captured by a Kinect sensor, which provides short-ranged semi-dense depth measurements.
For both training and testing, we simulate uniform depth point sampling to randomly select 500 depth samples for each image \& depth pair.
To neglect the image boundary area with missing depth measurements, we downsized each image into a size of $320 \times 240$, and then center-cropped it to be a size of $304 \times 228$.
We utilized $\sim$48K image pairs from the official training set and used handpicked 654 pairs among 1449 pairs of fully-labeled dataset for evaluating, following previous works.

\noindent{\textbf{KITTI Depth Completion dataset.}} The KITTI dataset~\cite{Geiger2012CVPR} is a driving scene dataset that was captured by stereo RGB cameras, stereo monochrome cameras, and a LiDAR sensor.
The KITTI Depth Completion dataset~\cite{Uhrig2017THREEDV} provides RGB image and 2D-projected LiDAR sensor measurement as a pair.
It utilizes 11 consecutive frames to generate denser depth map as a ground truth.
For training, we ignored the regions with no LiDAR measurements, by bottom-cropping the images with the size of $1216 \times 256$.
We used $\sim$93K image pairs for training, and official selected validation set of 1000 pairs for comparisons with other approaches.

\begin{figure*}[t]
\begin{center}
\includegraphics[width=\linewidth]{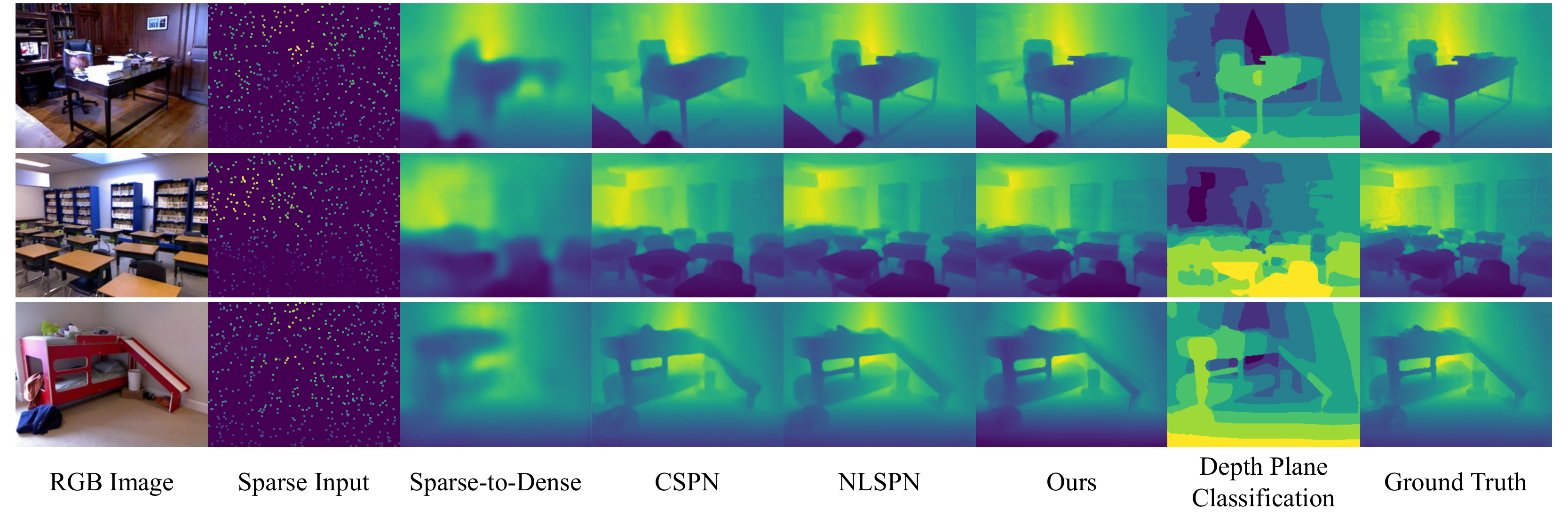}
\end{center}
\caption{\textbf{Qualitative depth completion results on the NYU Depth V2 dataset.} Depth plane classification results were shown in reversed color scheme for better visibility.}
\label{fig:nyu}
\vspace{-2mm}
\end{figure*}

For both datasets, we adopted ResNet-18~\cite{he2016resnet} model for our encoder, and used ADAM optimizer with learning rate of $1e^{-3}$, weight decay rate of $1e^{-6}$, and $\beta$ of $(0.9, 0.999)$.
The learning rate was reduced to its $20\%$ every 5 epochs, where the training was done for 20 epochs in total.
We set the batch size to 32 for the NYU Depth V2 dataset, and 8 for the KITTI Depth Completion dataset, using 4 NVIDIA RTX 2080 Ti GPUs, which take half a day and 3 days in training, respectively.

We set $d_1$ and $d_D$ as the minimum and the maximum depth values of the given sparse input samples.
We place the intermediate planes equally spaced depth-wise, thereby the value of $d_{\text{step}}$ is identical in every plane.
We chose the number of depth planes $D$ as 8 for the NYU Depth V2 dataset, and 64 for the KITTI Depth Completion dataset, regarding the maximum range difference between two datasets.
Comparison between other ways of choosing $d_p$s and $d_{\text{step}}$ will be shown in~\secref{sec:ablation}.

\begin{table}[t]
\begin{center}
\resizebox{0.45\textwidth}{!}{
\begin{tabular}{c|c|ccccc}
\hline
Method & \# params (M) & RMSE & REL & $\delta_1$ & $\delta_2$ & $\delta_3$ \\
\hline
Sparse-to-Dense~\cite{ma2018s2d} & 28.39 & 0.230 & 0.044 & 97.1 & 99.4 & 99.8 \\
DC-3coeff~\cite{imran2019dc} & 27.02 & 0.131 & \textbf{0.013} & 97.9 & 99.3 & 99.8 \\
DC-all~\cite{imran2019dc} & 27.02 & 0.118 & \textbf{0.013} & 97.8 & 99.4 & 99.9 \\
DeepLiDAR~\cite{qui2019deeplidar} & 53.44 & 0.115 & 0.022 & 99.3 & 99.9 & 100.0 \\
Ours & \textbf{14.34} & \textbf{0.104} & 0.014 & \textbf{99.4} & 99.9 & 100.0 \\
\hline
CSPN~\cite{cheng2018cspn} & 256.08 & 0.117 & 0.016 & 99.2 & 99.9 & 100.0 \\
Depth-Normal~\cite{xu2019depth} & 28.99 & 0.112 & 0.018 & 99.5 & 99.9 & 100.0 \\
NLSPN~\cite{park2020nlspn} & 25.84 & \textbf{0.092} & \textbf{0.012} & \textbf{99.6} & 99.9 & 100.0 \\
\hline
\end{tabular}}
\end{center}
\caption{\textbf{Depth completion evaluation results on the NYU Depth V2 dataset.} Algorithms on the lower block utilizes any kind of iterative post-refinement process. The metrics RMSE and REL are presented in meters (m). Numbers of parameters were driven from officially released codes.}
\label{table:nyu}
\vspace{-2mm}
\end{table}

\begin{figure*}[t]
\begin{center}
\includegraphics[width=\linewidth]{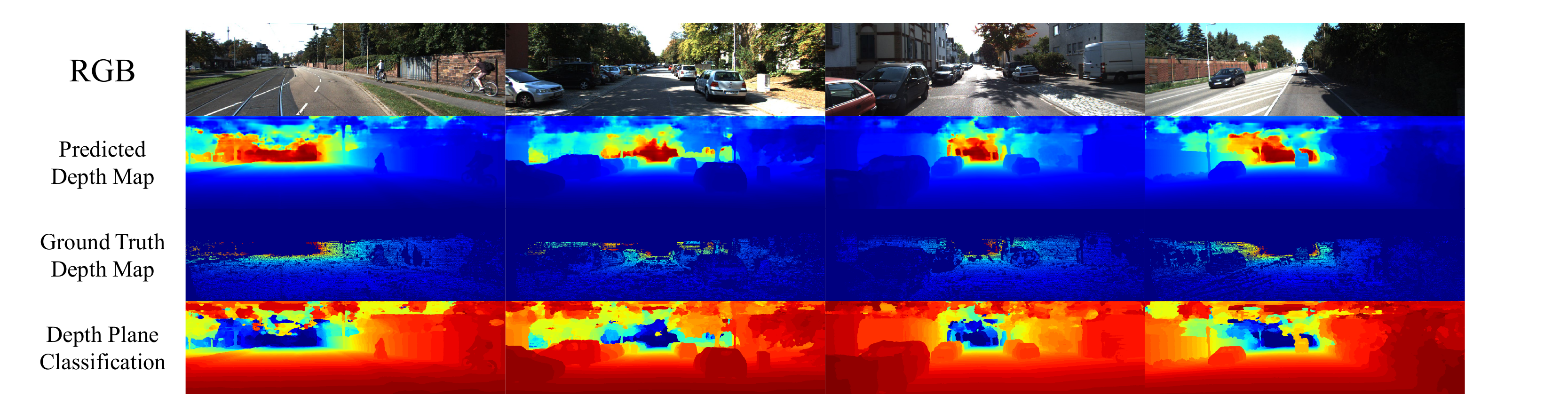}
\end{center}
\caption{\textbf{Qualitative depth completion results on the KITTI Depth Completion dataset.} Depth plane classification results were shown in reversed color scheme for better visibility.}
\label{fig:kitti}
\vspace{-2mm}
\end{figure*}

\subsection{Comparisons with state-of-the-art}

We use commonly used metrics for evaluation as following: root mean squared error (RMSE), mean absolute error (MAE), relative mean absolute error (REL), root mean squared error of the inverse depth (iRMSE), mean absolute error of the inverse depth (iMAE), and a percentage of predicted pixels with the relative error being smaller than $1.25^i$, $\delta_i$.

In~\tabref{table:nyu}, we present our quantitative evaluation result on the NYU Depth V2 dataset with others.
We categorized various depth completion algorithms into two categories, where the first one refers to the methods with no iterative post-processing refinement, and the other one means the opposite.
In~\tabref{table:nyu}, the upper block shows the previous approaches and ours, which are in the first category, and the lower block contains algorithms like CPSN~\cite{cheng2018cspn} and NLSPN~\cite{park2020nlspn}.

Among the methods with no additional iterative post-processing refinement, our result performs the best in most of the metrics.
More specifically, our depth completion network using PR representation outperforms Depth Coefficient (DC) algorithm by Imran~\etal~\cite{imran2019dc} in terms of RMSE and $\delta_i$s.
As was introduced in~\secref{sec:rw}, Depth Coefficient took similar strategy as ours, by defining multiple depth planes with discrete depth values.
They tried to predict depth coefficients using multi-class classification, which were then used for weighted summation in the final depth reconstruction.
Depth Coefficient used 80 channels of depth bases, and DC-all applied weighted summation using all of these 80 depth values, while DC-3coeff used only three adjacent depth values with the highest probability.
However, DC only performs classification, which is why we believe that they show reliable REL score, but it also requires sufficient number of depth channels to maintain certain level of performance.
On the other hand, our algorithm uses only 8 depth planes, yet combines the strength of classification and regression, thereby shows much better performance on other metrics.

\begin{table}[t]
\begin{center}
\resizebox{0.45\textwidth}{!}{
\begin{tabular}{c|c|cccc}
\hline
Method & Runtime (s) & RMSE & MAE & iRMSE & iMAE \\
\hline
Self-Supervised S2D~\cite{ma2019self} & 0.08 & 814.73 & 249.95 & 2.80 & 1.21 \\
DeepLiDAR~\cite{qui2019deeplidar} & 0.07 & \textbf{687.00} & 215.38 & 2.51 & 1.10 \\
Depth-Normal~\cite{xu2019depth} & 0.10 & 811.07 & 236.67 & 2.45 & 1.11 \\
DC-3coeff~\cite{imran2019dc} & 0.15 & 1212.00 & 241.00 & - & - \\
DC-all~\cite{imran2019dc} & 0.10 & 1061.00 & 252.00 & - & - \\
FuseNet~\cite{chen2019fusenet} & 0.09 & 785.00 & 217.00 & 2.36 & 1.08 \\
CSPN++~\cite{cheng2020cspn++} & 0.20 & 725.43 & 207.88 & - & - \\
Ours & \textbf{0.01} & 867.12 & \textbf{204.68} & \textbf{2.17} & \textbf{0.85} \\
\hline
\end{tabular}}
\end{center}
\caption{\textbf{Depth completion evaluation results on the KITTI Depth Completion validation dataset.} The metrics are presented in millimeters (mm). Running times were taken from the official KITTI benchmark website. Note that a few of the reported methods have different experimental environments.}
\label{table:kitti}
\vspace{-2mm}
\end{table}

We also report the number of network parameters of each algorithm, to analyze the efficiency of the approach.
Compared to CSPN~\cite{cheng2018cspn}, which uses ResNet-50 as encoder, and utilizes heavy and time-consuming iterative post-refinement module, our depth completion network using PR representation performs much better in every metrics, while having only 5$\sim$6\% of network parameters.
Also, our approach shows comparable result in comparison to NLSPN~\cite{park2020nlspn}, which is the state-of-the-art method, with $\sim$55\% of network parameters.
Overall, our result shows the second-best performance in most of the metrics with the smallest number of network parameters, that is 5$\sim$55\% of others.
This shows that our way of reformulating depth completion problem into a combination of depth plane classification and residual map regression is efficient and effective approach.

Qualitative results on the NYU Depth V2 dataset are also shown in~\figref{fig:nyu}.
Compared to NLSPN~\cite{park2020nlspn}, although we did not use any iterative post-refinement processing, our results show even better details in depth map estimations, as the wheels of the chair in the first example, or the details of the bookshelves in the second example.
This is due to our depth plane classification, which avoids depth mixing on object boundaries, and our context-guided probability volume filtering.
Also, since our decoder $R$ can learn normalized residual map plane-wise, it is easier to obtain more details by magnifying information between depth planes.

Evaluation results on the official KITTI Depth Completion validation dataset is shown in~\tabref{table:kitti}.
For the KITTI Depth Completion dataset, we also report running time of each algorithm, where the numbers were retrieved from official KITTI benchmark website.
Our algorithm has the shortest inference time, where it is 7 to 20 times faster, comparing to other algorithms.
Also, our results excel others in MAE, iRMSE, and iMAE metrics.
This again shows that our approach is efficient, in both network mobility and computational speed.
\figref{fig:kitti} shows qualitative results on depth completion and plane classification of our algorithm.

\subsection{Ablation Study}
\label{sec:ablation}

\begin{table}[t]
\begin{center}
\resizebox{0.45\textwidth}{!}{
\begin{tabular}{c|c|ccccc}
\hline
Category & Method & RMSE & REL & $\delta_1$ & $\delta_2$ & $\delta_3$ \\
\hline
\multirow{3}{*}{Depth Plane Setup} 
& UA & 0.113 & 0.016 & 99.3 & 99.9 & 100.0 \\
& DR & 0.114 & 0.016 & 99.3 & 99.9 & 100.0 \\
& DA & 0.117 & 0.017 & 99.1 & 99.8 & 99.9 \\
\hline
\multirow{3}{*}{Number of Planes} 
& 4 & 0.111 & 0.016 & \textbf{99.4} & 99.9 & 100.0 \\
& 16 & 0.105 & 0.015 & \textbf{99.4} & 99.9 & 100.0 \\
& 32 & 0.106 & 0.015 & \textbf{99.4} & 99.9 & 100.0 \\
\hline
$l(p)$ Filtering & No & 0.105 & 0.015 & \textbf{99.4} & 99.9 & 100.0 \\
\hline
Confidence-guided & No & 0.122 & 0.018 & 99.2 & 99.8 & 99.9 \\
\hline
\multicolumn{2}{c|}{Ours} & \textbf{0.104} & \textbf{0.014} & \textbf{99.4} & 99.9 & 100.0 \\
\hline
\end{tabular}}
\end{center}
\caption{\textbf{Ablation study on our design choices.} Other options beside selected category were set to default settings of `Ours'. Default settings are; UR depth plane setup, 8 planes, with $l(p)$ filtering, and with confidence-guided learning. The results were taken from the NYU Depth V2 dataset.}
\label{table:abl}
\vspace{-2mm}
\end{table}

\begin{figure*}[t]
\begin{center}
\includegraphics[width=\linewidth]{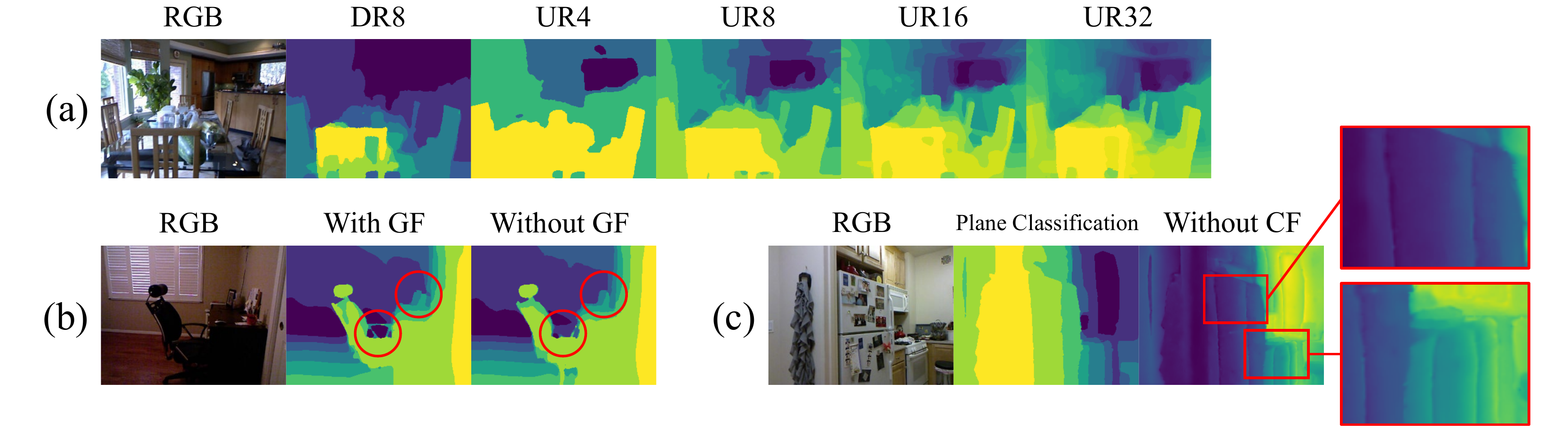}
\end{center}
\caption{\textbf{Ablation study on our design choices.} (a) Depth plane classification results on different depth planes setups. (b) Improvement on using channel-wise guided image filtering. (c) Failure case of not using confidence-guided residual learning.}
\label{fig:abl}
\end{figure*}

\noindent{\textbf{Setting the depth planes.}}
As mentioned in~\secref{sec:pr}, the values of plane depths and the number of planes can be set arbitrarily.
Our proposed way of determining the parameters is to set $d_1$ and $d_D$ according to the sparse input values, and to place intermediate planes equally spaced in between.
We name this setup UR, as in \textbf{U}niformly and \textbf{R}elatively set depth planes.

In~\tabref{table:abl}, we present three other setups, which are, UA (\textbf{U}niformly \& \textbf{A}bsolutely), DR (\textbf{D}isparity-wise \& \textbf{R}elatively), and DA (\textbf{D}isparity-wise \& \textbf{A}bsolutely).
For example, $d_1$ and $d_D$ were set to 0m and 10m in UA setup, regarding the overall depth range of the NYU Depth V2 dataset.
Some of the previous multiplane-related depth prediction approaches like DPSNet~\cite{im2018dpsnet} or DORN~\cite{fu2018deep} claimed that placing depth planes disparity-wise or in increasing log scale would boost the performance compared to uniformly-placed depth plane setup.
However, in the case of using PR representation for depth completion, the results show that UR and UA perform better than DR and DA.
Unlike DPSNet~\cite{im2018dpsnet} and DORN~\cite{fu2018deep}, which only focus on finding the best depth plane to place a pixel, our depth completion method using PR representation estimates residual map on top of the depth plane classification result.
Since we designed our predicted residual map to have a single channel, we believe that disparity-wise depth plane settings like DR and DA suffer from varying $d_{\text{step}}$ values, which hardens the residual regression.
Also, UA performs better than disparity-wise plane settings but worse than UR, since the distance that each plane should represent is broader.
On the other hand, DR preserves better object boundary information in close region, as shown in~\figref{fig:abl}-(a), because it placed more planes on the front.
We believe that the best way to setup depth planes in our PR representation would vary, depending on the task.

We also examine the influence of the number of planes.
The number of planes affects the ratio of classification and regression in our depth completion network.
As shown in~\tabref{table:abl}, our result with plane number of 8 performs the best, and the experiments with 16 and 32 number of depth planes show better results than the one with 4 depth planes.
Our intuition is that if the plane number is large, it lessens the burden for the decoder $R$.
However, setting more depth planes results in the increased difficulty of depth plane classification.
Moreover, it would make initial input on each plane to be very sparse, which will also make the plane classification part more challenging.

\noindent{\textbf{Probability volume filtering.}}
We show the effect of our probability volume filtering using channel-wise guided image filter.
Although our channel-wise guided filter do increase the quantitative evaluation result of the final predicted depth, it does not boost the performance dramatically, as can be seen in~\tabref{table:abl}.
However in~\figref{fig:abl}-(b), we can see that the depth plane classification result with our channel-wise guided image filter contains more detailed information on the object boundaries area, compared to the result with no filter applied.

\noindent{\textbf{Confidence-guided residual regression.}}
We examine on how confidence-guided adaptive residual learning improves the result.
As was explained in~\secref{sec:cf}, in order to make the reconstructed depth result more seamless on the depth plane boundary region, residual learning using depth plane classification-based confidence information is needed.
As presented in~\tabref{table:abl}, using confidence-guided learning improves very much of a performance.
Also, in~\figref{fig:abl}-(c), the result without confidence-guided learning finds it difficult to predict the right residual values at the boundary regions between two discrete planes, and therefore makes a discontinuity in the final predicted depth map.

\section{Conclusion}

In this paper, we addressed the main difficulties in previous depth completion algorithms, which are depth mixing on object boundaries, heavy network computation, and slow inference.
We proposed a novel way to solve depth completion, by reformulating depth regression problem into a combination of depth plane classification and residual regression.
We introduced Plane-Residual (PR) representation as well, where we represent a depth pixel as two separate values as in $(p, r)$, where $p$ is the closest discrete depth plane, and $r$ is the distance from the plane.
We showed competitive results with lighter and faster computation using our approach, thereby verified our idea and design choices.
We also believe that PR representation can be utilized in various depth-related problems like monocular depth estimation and multi-view stereo matching.

{\small
\bibliographystyle{ieee_fullname}
\bibliography{egbib}
}

\end{document}